\documentclass[conference]{IEEEtran}
\IEEEoverridecommandlockouts
\usepackage{cite}
\usepackage{amsmath,amssymb,amsfonts}
\usepackage{algorithmic}
\usepackage{graphicx}
\usepackage{textcomp}
\usepackage{hyperref}
\usepackage{xcolor}
\def\BibTeX{{\rm B\kern-.05em{\sc i\kern-.025em b}\kern-.08em
    T\kern-.1667em\lower.7ex\hbox{E}\kern-.125emX}}

\makeatletter
\newcommand{\linebreakand}{%
  \end{@IEEEauthorhalign}
  \hfill\mbox{}\par
  \mbox{}\hfill\begin{@IEEEauthorhalign}
}
\makeatother

\begin{document}

\title{Multi-perspective Imbalance-Conscious 6G Beamforming Optimization and Performance }

%Faculty of Business \& IT
%Ontario Tech University
%Joint Admissions \& Matn. Board
%School of Physics, Engineering \& Computer Science
%SPECS

\author{%
\IEEEauthorblockN{Chukwunonso Henry Nwokoye}
\IEEEauthorblockA{\textit{Liberal Arts \& Professional Studies} \\
\textit{York University,}\\
Toronto, Canada \\
0000-0002-2534-3734}
\and
\IEEEauthorblockN{Blessing Oluchi Iloka}
\IEEEauthorblockA{\textit{Physics, Engineering \& Computer Science} \\
\textit{University of Hertfordshire}\\
England, United Kingdom \\
Boluchi23@gmail.com}
\and
\IEEEauthorblockN{Chikwue V. Umeugoji}
\IEEEauthorblockA{\textit{IT Services,} \\
\textit{Joint Admissions \& Matn. Board}\\
Lagos, Nigeria \\
Umeugoji.chikwue@gmail.com}

\linebreakand

\IEEEauthorblockN{Christopher Anene Egemba}
\IEEEauthorblockA{\textit{AI Product Development} \\
\textit{Lightenet Technologies Ltd.}\\
United Kingdom \\
0009-0007-2228-8243}
\and
\IEEEauthorblockN{Nnenna D. Duroha}
\IEEEauthorblockA{\textit{Physics, Engineering \& Computer Science} \\
\textit{University of Hertfordshire}\\
England, United Kingdom\\
nnennadaisy@yahoo.com}
}
\vspace{-9pt}

\maketitle

\begin{abstract}
The study presents a systematic machine learning (ML) study of 6G-IoT beamforming optimization (6GBO) using supervised and unsupervised approaches. We compared the predictive power of network, environmental, device, and vision feature groups for 6GBO. Additionally, it addressed other unsupervised perspectives that can enhance 6GBO, including clustering network scenarios using methods such as K-means, DBSCAN, and hierarchical clustering. Several imbalance-aware experiments revealed that network features possess better prediction power than device, environmental and vision feature groups, as evidenced by their recall, F1-score and ROC-AUC values. For unsupervised ML exploration (assessed using  Elbow, Silhouette score and Davies Bouldin Index methods), the results indicate that the deployment environment and type of device primarily influence clustering, rather than mobility-based attributes. Furthermore, the explainability analysis showed that bandwidth, IoT sensors, and mobility possess higher global feature importance across the feature groups. In the future, we would apply deep and reinforcement learning techniques to predict throughput/latency or to optimize rewards determined by performance indicators like SNR enhancement.

\end{abstract}

\begin{IEEEkeywords}
Sixth Generation beamforming optimization (6GBO), Internet of Things (IoT), machine learning (ML), imbalance handling, Clustering, eXplainable AI (XAI)
\end{IEEEkeywords}

\vspace{-8pt}

\section{Introduction} 
The integration of the Internet of Things (IoT) with sixth-generation (6G) networks is poised to revolutionize global connectivity by enabling intelligent, ultra-reliable, and low-latency communication across diverse sectors, including industrial automation, smart cities, and autonomous systems \cite{QADIR2023296}. As 6G-IoT environments grow increasingly multifaceted \cite{AKBAR2025104040}, they face substantial challenges in dynamic situations where the allocation of resources and reliable communication are hindered by user mobility, network topologies, and rapid channel variations. In such circumstances, static beamforming techniques are insufficient; instead, smart and adaptive beamforming approaches capable of instantaneously responding to changing signal contexts are necessary. Beamforming is a signal processing method employed in sensor or antenna arrays to guide signal transmission or reception by modifying the amplitude and phase of each component, thereby improving signal quality and minimizing interference \cite{Benesty2021}. This phenomenon boosts signal strength, mitigates interference, and improves the overall performance of the network \cite{saad2019vision, giordani2020toward}.

With the advancement of wireless communication networks in the direction of 6G \cite{AKBAR2025104040}, beamforming has become a dependable method for attaining higher data rates, improved spectral efficiencies, and ultra-reliable low-latency communication (URLLC) \cite{duong2023ultra}. Beamforming enhances transfer and reception by concentrating signal energies on certain devices or users, which is crucial in high-frequency bands like THz and mmWave, where signal reduction is unavoidable \cite{heath2016overview, saad2019vision}. However, conventional static beamforming techniques face considerable obstacles in dynamic 6G-IoT settings marked by user movement, high device density, and rapidly changing channel scenarios that demand real-time flexibility. Adaptive beamforming approaches, which are supported by machine learning (ML), have shown significant potential in overcoming these challenges \cite{elbir2023twenty}. ML models can actively assess real-time data, predict ideal beam trajectories, and independently adjust transmission techniques to enhance reliability and performance \cite{huang2018deep}.

We believe there is a need to move beyond monolithic modeling of beamforming success by isolating distinct feature domains—such as network-level parameters, environmental context, or device characteristics—and comparing their relative predictive power. Such approaches allow researchers to ask systems-level questions like “Which factor category (network, environment, device, vision) most strongly determines beamforming success?” or “Are environmental conditions more critical than network configurations?” From a 6G resource-allocation and adaptive beamforming perspective, this analysis is highly insightful: it enables network designers to prioritize measurement and processing investments toward the domains that drive performance, rather than assuming all features carry equal significance. Considering these questions, this study aims to conduct a comparative analysis of 6G beamforming optimization (6GBO) using network, environmental, device and vision factors with ML classifiers. %Additionally, this study seeks to understand how these factors impact beamforming optimization. 

\section{RELATED WORK}
\label{lit_rew}
Prior research used supervised and unsupervised ML methods in conventional wireless systems for channel prediction and dynamic beamforming. Zhou et al. \cite{ZHOU2024} conducted an examination of ML solutions in IoT-supported wireless networks, highlighting the use of decision trees and neural networks for link adaptation and interference reduction. Qin et al. \cite{qin2019deep} showed that deep learning algorithms can effectively simulate signal processing at the physical layer, offering reliable solutions for channel feedback and beam option selection in variable situations. Reinforcement learning (RL) has garnered interest for its capacity to manage sequential decision-making processes in beamforming. Ye et al. \cite{ye2019deep} used deep RL to enhance resource allocation in vehicular networks, demonstrating its efficacy in adjusting beam directions within quickly evolving topologies.

The integration of beamforming optimization and ML \cite{Klautau2018} is facilitating the development of scalable, intelligent, and adaptable wireless networks, thereby establishing a basis for the ongoing expansion and robustness of next-generation communication infrastructure \cite{brilhante2023literature}. The incorporation of ML into wireless communication networks has garnered considerable attention owing to its capacity to tackle the issues posed by dynamic settings, high mobility, and optimal resource allocation. Specifically, ML-driven beamforming approaches have demonstrated potential for performance improvement in next-generation networks, including 5G \cite{Klautau2018}, and 6G \cite{brilhante2023literature}, as well as intelligent reflecting surface communication technologies.

Our study is aimed at understanding the impact of these factors on beamforming optimization using machine learning models: Logistic Regression (LR), XGBoost (XGB), Random Forest (RF), Gradient Boosting (GB), Support Vector Machine (SVM) using a radial basis function (RBF) kernel, Multilayer Perceptron (MLP), AdaBoost (AB), Gradient Boosting (GB), LightGBM, Extra Trees, and K-Nearest Neighbors (KNN). In contrast to previous GitHub implementations that utilize the entire dataset \cite{sharif_2025}, encompassing derived performance metrics, our study intentionally excludes the pre-decision feature groups (FGs)—environmental, network, device, and vision—to evaluate their individual predictive efficacy for beamforming optimization success. %We omit outcome measures from the set of predictors to prevent leakage. This facilitates an accurate evaluation of which visible, deployable metrics can inform real-time beamforming regulation and resource distribution. 

\section{Methodology}
\label{met}

We adopted the ML development life cycle, which includes data collection; preprocessing and feature engineering (normalization, handling missing values, and encoding); dataset splitting; modeling (using classifiers); evaluation; and results. The 6G IoT intelligent management dataset \cite{ziya2024} used for the study was accessed from Kaggle. Input features for our study include network (frequency (GHz), transmit power (dBm), bandwidth (MHz), codebook size, and interference level (dB)); environment (obstacle density, mobility (m/s), and environment type (outdoor)); and device (number of antennas and device type (IoT sensor and smartphone)). These input features are available before the occurrence of optimization. While running the experiments, the features were defined as groups, i.e., network parameters (NP), environmental factors (EF), device characteristics (DC), vision attributes (VA), and all features (AF). Note that VAs are Scale-Invariant Feature Transform (SIFT) keypoints, which are also part of the 6G IoT intelligent management dataset \cite{ziya2024}. Essentially, the target variable is the column called "optimized", which contains 0s and 1s for unsuccessful and successful 6GBO. To evaluate the performance of the models, we employed the following metrics: precision (Pre), recall (Rec), F1, ROC-AUC, and PR-AUC.

Prior to model training, missing values were replaced with the median and standardized, whereas categorical variables were substituted with the mode and subjected to one-hot encoding. Boolean attributes were transformed into floating-point representations. Additionally, a stratified division of 80:20 for training and testing was implemented with a constant random seed (42) to maintain class distributions. On class balance, the positive rate is 0.168 (i.e., 168 positives out of 1000). In every aspect of preprocessing—such as feature normalization, median imputation, one-hot encoding, Boolean feature transformation, and SMOTE oversampling (when applicable)—were integrated into an imblearn pipeline and implemented separately within each cross-validation training fold, consequently averting the issue of data leakage. 
%GridSearchCV was incorporated into the classification workflow subsequent to dataset preprocessing, and the models were assessed utilizing the reserved test set.

In our experiments, we utilized a 5-fold cross-validation (CV) technique for out-of-fold (OOF) threshold tuning (TT). By implication, the training dataset was partitioned into 5 stratified subsets, with 4 folds allocated for training and one fold designated for validation. The method was continuously repeated till every fold had functioned once as the validation set. The aggregated OOF prediction probabilities were utilized to establish the ideal classification threshold, which improved the F1-score in a situation of significant class imbalance. This approach is essentially distinct from GridSearchCV (GSCV), which was utilized largely for the optimization of hyperparameters (Table \ref{tab:gridsearch}) and the selection of models by determining the optimal parameter combinations according to cross-validation scoring measures.

\vspace{-9pt}
%\vspace{-0.5pt}

\begin{table}[htbp]
\caption{GridSearchCV Hyperparameter Search Space}
\label{tab:gridsearch}
\centering
\scriptsize
%\begin{tabular}{|l|l|}
\begin{tabular}{|l|p{7cm}|}
\hline
\textbf{Classifier} & \textbf{Hyperparameter Search Space} \\
\hline
LR & $C \in \{0.01, 0.1, 1.0\}$ \\
\hline
RF & $n_{\text{estimators}} \in \{100,200\}$, $max\_depth \in \{\text{None},10\}$ \\
%& \\
\hline
XGB & $n_{\text{estimators}} \in \{100,200\}$, $max\_depth \in \{3,6\}$,\\
%& \\
& $learning\_rate \in \{0.1,0.01\}$ \\
\hline
GB & $n_{\text{estimators}} \in \{100,200\}$, $learning\_rate \in \{0.1,0.01\}$ \\
%& \\
\hline
AdaBoost & $n_{\text{estimators}} \in \{50,100\}$, $learning\_rate \in \{1.0,0.1\}$ \\
%& \\
\hline
SVM & $C \in \{0.1,1.0\}$, $\gamma \in \{\text{scale},\text{auto}\}$ \\
%& \\
\hline
MLP & Hidden layers $\in \{(64,32),(128,64)\}$, $\alpha \in \{10^{-4},10^{-3}\}$\\
%& \\
\hline
\end{tabular}
\end{table}

Table \ref{tab:ablation_protocol} presents the ablation study protocol used to evaluate the effect of feature leakage on 6GBO classification. Therein, Model 2 input features include Network, Environment, Device and Vision (pre-decision only); Model 2 input features include performance metrics only, and Model 3 features include pre-decision and performance metrics (outcomes). Model 2 and 3 was implemented to illustrate the significant jump in evaluation metrics (accuracy/AUC), thus demonstrating target leakage. Other columns in Table \ref{tab:ablation_protocol} present methods applied in order to address class imbalance.

\begin{table}[htbp]
\caption{Ablation Study Protocol}
\begin{center}

\scriptsize

\begin{tabular}{|c|c|c|c|c|}
\hline
\textbf{Model} & \textbf{Feature Configuration} & \textbf{SMOTE} & \textbf{TT} & \textbf{GSCV} \\
\hline

1 &
Pre-decision
(NP, EF, DC, VA) &
Yes &
Yes  &
No \\
\hline

1 &
Pre-decision
(NP, EF, DC, VA) &
Yes &
No  &
Yes \\
\hline

1-Ensembles &
Pre-decision
(NP, EF, DC, VA)  &
Yes &
Yes  &
No \\
\hline

2 &
Performance Metrics Only &
Yes &
No  &
No \\
\hline

3 &
Model 1 + Performance Metrics &
No &
No  &
No \\
\hline
\end{tabular}
\label{tab:ablation_protocol}
\end{center}
\end{table}

\vspace{-9pt}

\section{Results \& Discussion}
\label{res}
%\section{Discussion}
%\label{sec:discussion}

%Following the ML life cycle and specifically conducting preprocessing steps such as checking for missing values, scaling using the standard scaler, and encoding (one-hot) for categorical values (Environment: Outdoor; Device Type: IoT Sensor, Smartphone). 

The initial experiment reveals that class imbalance dominates the observed performance. Recall that the dataset’s positive rate (optimized = 1) is small, so many models converge to predicting the majority (non-optimized) class. As a consequence, overall accuracy is high ($\approx$0.83) but provides a misleading impression of success: several classifiers (notably those trained only on device-level features) report precision = 0, recall = 0, and F1 = 0, indicating that they never detect any positive examples. This behavior matches the majority-class baseline (predicting all samples as negative) and shows that accuracy alone is not an informative metric in this setting. Results of our experiments are available at the following GitHub link: https://github.com/ChiNonsoHenry16/Multi-perspective-Imbalace-Aware-6G-Beatforming-Optimization. 

Through preliminary experiments, a more discriminating view is obtained from the class-sensitive metrics and the probabilistic ranking metrics (ROC-AUC). ROC-AUC values across the experiments lie near random to weakly informative (approximately 0.44–0.57), with the best AUCs reaching roughly 0.56–0.57 for a few models trained on environmental or network features. Correspondingly, only a handful of models (e.g., RF and XGBoost on network/environmental features) produce non-zero recall and modest F1 scores (examples: Environmental RF F1 $\approx$ 0.21, Environmental XGBoost F1 $\approx$ 0.15, and Network XGBoost F1 $\approx$ 0.08). These non-zero values indicate there is some discriminative signal in network and environmental features, but it is weak and insufficient for reliable detection with the current data and training setup. 

Due to the significant dataset class imbalance (positive class of 16.8\%), the evaluation of model performance predominantly relied on Recall, F1-score, ROC-AUC, and PR-AUC, as these metrics offer a more nuanced evaluation of minority-class identification compared to classification accuracy. Accuracy is conveyed as a supplementary measure for thoroughness.

%This feature configuration is corroborated by experimental-based verification. We expect moderate ROC-AUC and F1-scores owing to difficulties arising from class imbalance, underscoring its efficacy for dynamic and real-world 6G beamforming optimization.  

\subsection{Model 1: Pre-optimization Factors Only}
Model 1 presents a deployable prediction pipeline that employs pre-decision data. Here, the classifiers were used on just the network, environmental, device and vision parameters alone, without the addition of the performance metrics. They are as follows: NP (Frequency, Transmit Power, Bandwidth, Codebook Size, Interference Level; EF (Obstacle Density, Mobility, Environment-Outdoor); DC (Number of Antennas, Device Type-IoT Sensor, Device Type-Smartphone); and vision attributes (VA) (SIFT). In other words, these parameters are derived pre-optimization and exclude the following: Beamforming Gain, Latency, Energy Consumption, Throughput, Beam Training Time, SNR Improvement, Processing Time, and Memory Usage. 

%They are as follows: NP (Frequency (GHz), Transmit Power (dBm), Bandwidth (MHz), Codebook Size, Interference Level (dB)); EF (Obstacle Density, Mobility (m/s), Environment-Outdoor); DC (Number of Antennas, Device Type-IoT Sensor, Device Type-Smartphone); and vision attributes (VA) (SIFT). In other words, these parameters are derived pre-optimization and exclude the following: Beamforming Gain (dB), Latency (ms), Energy Consumption (kWh/Gb), Throughput (Mbps), Beam Training Time (s), SNR Improvement (dB), Processing Time (ms), and Memory Usage (MB). 

Tables \ref{tab:default_threshold} and \ref{tab:tuned_threshold} display the best-performing (BP) ML models for the classification of 6GBO at a decision threshold (of 0.5) and the tuned (adjusted) threshold conditions, respectively. The code implementation employed libraries for preprocessing, encoding (one-hot), SMOTE for handling imbalance, probability calibration, and threshold refinement to tackle the significant class imbalance. %The dataset had merely 168 optimized samples from 1000 data points, yielding a positive rate of 16.8\%, which considerably impacted the evaluation of classification performance.

At a 0.5 threshold, the network and environmental groups demonstrated a comparably better performance, with AdaBoost and XGBoost attaining the highest F1 values of 0.327 and 0.324, respectively. After TT, recall significantly enhanced every FG, with some models attaining recall values around or equal to 1.0. LightGBM achieved the highest F1-score of 0.301 for NP, while SVM (RBF) attained the same score of 0.301 for vision attributes. The results demonstrate that network attributes offer strong predictive signals for 6GBO, whereas environmental, vision, and device features supply supplementary data for ML classification. No individual classifier consistently excelled throughout all feature domains, indicating significant disparities in the fundamental data distributions across the feature spaces.

Numerous significant insights arise from Tables \ref{tab:default_threshold} and \ref{tab:tuned_threshold}. Firstly, TT significantly enhanced recall among all FG, affirming the significance of imbalance-conscious classification for 6GBO predictions. Secondly, NP consistently demonstrated a comparably better overall performance, whereas environmental and device features yielded identical F1-scores, suggesting that contextual and device-specific information significantly contributes to beamforming efficacy. Thirdly, the utilization of all features did not substantially exceed the performance of network parameters alone, indicating that supplementary feature groups offer minimal additional predictive capability. Ultimately, various feature groups preferred distinct ML techniques, with LightGBM, XGBoost, LR, MLP, and SVM (RBF) identified as the most effective models for specific feature domains, underscoring the varied characteristics of 6GBO prediction tasks.

\vspace{-8pt}

\begin{table}[htbp]
\caption{Best-Performing Models at Default Threshold (\(0.5\))}
\label{tab:default_threshold}
\centering
\scriptsize
\begin{tabular}{|l|l|c|c|c|c|}
\hline
\textbf{Feature Group (FG)} & \textbf{Model} & \textbf{Acc.} & \textbf{Prec.} & \textbf{Recall} & \textbf{F1} \\
\hline
All Features (AF) & AdaBoost & 0.645 & 0.224 & 0.441 & 0.297 \\
Network Parameters (NP) & AdaBoost & 0.505 & 0.212 & 0.706 & 0.327 \\
Environmental Factors (EF) & XGBoost & 0.520 & 0.213 & 0.676 & 0.324 \\
Device Characteristics (DC) & XGBoost & 0.170 & 0.170 & 1.000 & 0.291 \\
Vision Attributes (VA) & AdaBoost & 0.285 & 0.166 & 0.794 & 0.274 \\
\hline
\end{tabular}
\end{table}

%\vspace{-10pt}

\begin{table}[htbp]
\caption{Best Performing Models After Threshold Tuning}
\label{tab:tuned_threshold}
\centering
\scriptsize
\begin{tabular}{|l|l|c|c|c|c|c|c|}
\hline
\textbf{FG} & \textbf{Model} & \textbf{Acc} & \textbf{Prec} & \textbf{Rec} & \textbf{F1} & \textbf{ROC-AUC} & \textbf{PR-AUC} \\
\hline
AF & MLP & 0.205 & 0.176 & 1.000 & 0.300 & 0.533 & 0.195 \\
NP & LightGBM & 0.210 & 0.177 & 1.000 & 0.301 & 0.562 & 0.202 \\
EF & XGBoost & 0.180 & 0.172 & 1.000 & 0.293 & 0.571 & 0.203 \\
DC & LR & 0.200 & 0.172 & 0.971 & 0.292 & 0.533 & 0.184 \\
VA & SVM & 0.280 & 0.180 & 0.912 & 0.301 & 0.466 & 0.164 \\
\hline
\end{tabular}
\end{table}

Despite some classifiers attaining elevated accuracy@0.5 levels between 0.70 and 0.83, a majority of these models demonstrated significantly deficient recall and F1-score results owing to the pronounced class imbalance within the dataset. %As a result, the models are predominantly trained to forecast largely the non-optimized category. Consequently, the study emphasizes imbalance-sensitive metrics, including F1-score, recall, ROC-AUC, and PR-AUC, in determining the most functionally successful models for beamforming optimization.

Table \ref{tab:gridsearch_best_models} displays the top-performing models using SMOTE and optimized by GridSearchCV, as determined by F1-score. Across all groups, DC attained the best performance, with AdaBoost attaining the highest F1 score of 0.307 and an ROC-AUC of 0.564. The SVM (RBF) classification algorithm attained the highest F1 score of 0.256 for NP, while LR excelled for EF with a 0.266 F1 score. Considering the VA category (with SIFT keypoints), AdaBoost attained the highest F1-score of 0.240. These findings indicate that device-related characteristics yield a better predictive signal for 6GBO, but environmental, network, and VA-based parameters demonstrate moderate predictive efficacy when assessed in isolation. While the highest standalone F1 value in the GridSearchCV studies was attained by AdaBoost utilizing device-related characteristics (F1 = 0.307), network attributes consistently yielded superior ROC-AUC and PR-AUC values across many experimental configurations. Thus, NP-related properties seem to yield better discriminatory information for predicting 6GBO, whereas device characteristics present a desirable precision-recall equilibrium at specific classification levels.

\vspace{-9pt}

\begin{table}[htbp]
\caption{Best GridSearchCV-Tuned Models by Feature Group}
\label{tab:gridsearch_best_models}
%\centering
\scriptsize
\begin{tabular}{|l|l|c|c|c|c|c|}
\hline
\textbf{FG} & \textbf{Best Model} & \textbf{Accuracy} & \textbf{Precision} & \textbf{Recall} & \textbf{F1 } & \textbf{ROC-AUC} \\
\hline
NP & SVM (RBF) & 0.457 & 0.166 & 0.560 & 0.256 & 0.537 \\
\hline
EF & LR & 0.540 & 0.181 & 0.500 & 0.266 & 0.492 \\
\hline
DC & AdaBoost & 0.563 & 0.209 & 0.580 & 0.307 & 0.564 \\
\hline
VA & AdaBoost & 0.493 & 0.160 & 0.480 & 0.240 & 0.484 \\
\hline
\end{tabular}
\end{table}

\vspace{-7pt}

\subsection{Model 1: Ensembles}

Here, we employed voting ensemble (VE) and stacking ensemble (SE) algorithms for 6GBO classification. Tables \ref{tab:Ensemble@0.5} and \ref{tab:Ensemble@Tuned} contains several classification results following preprocessing, SMOTE-based imbalance reduction, and OOFTT. At the default TT of 0.5, numerous models attained commendable accuracy levels, especially within the AF and NP categories; yet, recall and F1-scores were rather low, signifying inadequate identification of the minority class due to the imbalance. Following TT, recall significantly enhanced across the majority of feature groups, with numerous models attaining recall values around or equal to 1.000. 

The network FG exhibited the best predictive performance, with the VE attaining the largest ROC-AUC and PR-AUC of 0.585 and 0.224, respectively. Also, the SE for network parameters recorded the highest optimized F1-score (0.302). The environment and device features yielded a somewhat moderate performance, whereas the vision (SIFT) attributes exhibited relatively poor discriminative ability. The findings indicate that network-related characteristics are the primary contributors to the prediction of adaptive and dynamic 6GBO. Additionally, beyond the ML-based 5G beam selection analysis in Klautau, et al. \cite{Klautau2018}, the ensembles (SE and VE) in our study achieved higher accuracies (Table \ref{tab:Ensemble@0.5}). However, it is noteworthy that both studies were conducted using different datasets. Interestingly, the device attributes achieved the strongest F1-score (0.307) when GridSearchCV was employed. 

%our study included ensembles with higher accuracies. 
%CLASSIFICATION RESULTS AT DEFAULT THRESHOLD (0.5)
%CR@DT

%\begin{table*}[htbp]
\begin{table}[htbp]
\caption{Classification Results at Default Threshold (\(0.5\))}

\begin{center}

\scriptsize

\begin{tabular}{|l|l|c|c|c|c|}
\hline
\textbf{FG} & \textbf{Model} & \textbf{Acc} & \textbf{Pre} & \textbf{Rec} & \textbf{F1} \\
\hline

AF & VE & 0.780 & 0.222 & 0.118 & 0.154 \\
AF & SE & 0.810 & 0.167 & 0.029 & 0.050 \\
\hline

DC & SE & 0.530 & 0.194 & 0.559 & 0.288 \\
DC & VE & 0.170 & 0.170 & 1.000 & 0.291 \\
\hline

EF & SE & 0.690 & 0.150 & 0.176 & 0.162 \\
EF & VE & 0.580 & 0.143 & 0.294 & 0.192 \\
\hline

NP & SE & 0.800 & 0.313 & 0.147 & 0.200 \\
NP & VE & 0.745 & 0.282 & 0.324 & 0.301 \\
\hline

VA & VE & 0.400 & 0.123 & 0.412 & 0.189 \\
VA & SE & 0.535 & 0.127 & 0.294 & 0.177 \\
\hline

\end{tabular}
\label{tab:Ensemble@0.5}
\end{center}
\end{table}

\vspace{-20pt}

\begin{table}[htbp]
\caption{6GBO Classification Results after Threshold Tuning}
\begin{center}

\scriptsize

\begin{tabular}{|l|l|c|c|c|c|c|c|}
\hline
\textbf{FG} & \textbf{Model} & \textbf{Acc} & \textbf{Pre} & \textbf{Rec} & \textbf{F1} & \textbf{ROC-AUC} & \textbf{PR-AUC} \\
\hline

AF & VE & 0.170 & 0.170 & 1.000 & 0.291 & 0.476 & 0.181 \\
AF & SE & 0.165 & 0.162 & 0.941 & 0.277 & 0.421 & 0.156 \\
\hline

DC & SE & 0.170 & 0.170 & 1.000 & 0.291 & 0.494 & 0.174 \\
DC & VE & 0.170 & 0.170 & 1.000 & 0.291 & 0.467 & 0.168 \\
\hline

EF & SE & 0.190 & 0.170 & 0.971 & 0.289 & 0.504 & 0.171 \\
EF & VE & 0.175 & 0.168 & 0.971 & 0.286 & 0.517 & 0.187 \\
\hline

NP & SE & 0.445 & 0.192 & 0.706 & 0.302 & 0.577 & 0.216 \\
NP & VE & 0.170 & 0.170 & 1.000 & 0.291 & 0.585 & 0.224 \\
\hline

VA & VE & 0.210 & 0.167 & 0.912 & 0.282 & 0.411 & 0.153 \\
VA & SE & 0.235 & 0.164 & 0.853 & 0.275 & 0.408 & 0.149 \\
\hline

\end{tabular}
\label{tab:Ensemble@Tuned}
\end{center}
\end{table}

\vspace{-4pt}

\subsection{Model 2: Performance Metrics Alone}
The objective of Model 2 is to assess the extent to which post-hoc outcome factors independently forecast the optimized label. This model was implemented as a benchmark for validating leakage. The pre-optimization data were excluded in the experiment conducted here. The performance metrics include the following: Beamforming Gain, Latency, Energy Consumption, Throughput, Beam Training Time, SNR Improvement, Processing Time, and Memory Usage. 

%It is expected that the model will show exceptional classification performance, which confirms that performance measures exhibit a strong correlation with the optimized label; hence, it is not recommended to be utilized in deployable real-world prediction-focused beamforming systems.

Table \ref{tab:performance_metrics_only} summarizes the 6GBO findings derived just from the above-mentioned performance metrics. The evaluations included preprocessing and SMOTE-based imbalance management to assess the predictive strength of the post-optimization outcome on the optimized label. The findings demonstrated good classification performance across almost all ML models, with GB, AdaBoost, DT, HG, XGBoost, and LightGBM attaining excellent scores of 1.000 for all evaluation metrics. Likewise, RF attained nearly flawless performance, exhibiting an accuracy and ROC-AUC of 0.995 and 1.000, respectively. Even relatively worse models like KNN and logistic regression yielded impressive ROC-AUC values of 0.935 and 0.944, respectively.

The findings indicate a good correlation between the performance measures as well as the target label, thereby providing significant post-hoc information regarding beamforming performance. %Thus, the nearly flawless prediction performance demonstrated in this experiment suggests target leakage, wherein outcome-dependent parameters implicitly reflect the optimization result. Although these models may seem exceptionally precise, they are unsuitable for practical implementation since numerous performance metrics are only accessible post-beamforming optimization.
Consequently, employing these factors for real-time forecasting will artificially enhance classification performance and inadequately represent actual decision-making scenarios in adaptive 6GBO. To a large extent, our results thus confirm the necessity of limiting deployable prediction pipelines of 6GBO to pre-decision (pre-optimization) data. This distinction is essential for guaranteeing equitable evaluation and accurate assessment of intelligent 6GBO schemes in emerging wireless communication settings.

\begin{table}[htbp]
\caption{Best Performing Models Using Performance Metrics Only}
\begin{center}

\scriptsize

\begin{tabular}{|l|c|c|c|c|}
\hline
\textbf{Model} & \textbf{Accuracy} & \textbf{F1-Score} & \textbf{ROC-AUC} & \textbf{PR-AUC} \\
\hline

GB & 1.000 & 1.000 & 1.000 & 1.000 \\
\hline

AdaBoost & 1.000 & 1.000 & 1.000 & 1.000 \\
\hline

Decision Tree & 1.000 & 1.000 & 1.000 & 1.000 \\
\hline

HGB & 1.000 & 1.000 & 1.000 & 1.000 \\
\hline

XGBoost & 1.000 & 1.000 & 1.000 & 1.000 \\
\hline

LightGBM & 1.000 & 1.000 & 1.000 & 1.000 \\
\hline

RF & 0.995 & 0.985 & 1.000 & 1.000 \\
\hline

Extra Trees & 0.970 & 0.906 & 0.996 & 0.979 \\
\hline

MLP & 0.950 & 0.853 & 0.983 & 0.931 \\
\hline

\end{tabular}

\label{tab:performance_metrics_only}

\end{center}
\end{table}

\vspace{-9pt}

\subsection{Model 3: All Features (Model 1 + Model 2)}

The objective of Model 3 is to assess the impact of integrating pre-decision attributes with post-optimization result metrics to examine the effects of target leakage. Table \ref{tab:all_features_no_smote} displays the classification outcomes achieved with the entire feature set, excluding SMOTE balancing and threshold adjustment. Multiple ensemble models, notably GB, AdaBoost, DT, XGBoost, HGB, and LightGBM, attained excellent performance for every evaluation metric (1.000). Likewise, RF attained an exceptional result with accuracy and ROC-AUC of 0.985 and 1.000, respectively. Classical ML models, including LR and SVM (RBF), exhibited good prediction performance, attaining ROC-AUC values over 0.93.

While these results first imply a highly superior prediction of 6GBO, the perfect performance clearly suggests the existence of target leakage inside the feature space. The incorporation of post-hoc performance indicators furnishes the models with outcome-related data that is inherently linked to the optimized label. Our decision to remove SMOTE balancing and TT is to clearly show that this model achieved perfect performance without these imbalance-aware approaches. Thus, the classifiers are proficiently learning optimization results instead of forecasting circumstances based on actionable pre-decision parameters.

%This model is similar to Sharif's GitHub implementations \cite{sharif_2025}, and as expected, it generated a significant enhancement in accuracy, recall, ROC-AUC, and F1-score, as performance metrics inherently reflect the efficacy of optimization efforts. The model grossly exhibits performance inflation induced by leakage.

Like Sharif's implementations \cite{sharif_2025}, this model generated a significant enhancement in all metrics. Therefore, the findings also underscore the necessity of meticulously distinguishing pre-decision factors from post-optimization performance measurements in the development of smart beamforming systems. The entire feature configuration yields remarkably high classification values; nevertheless, these outcomes do not accurately represent practical deployment scenarios, as several included metrics are accessible only post-beamforming optimization. This discovery substantiates the requirement for the ablation technique presented above and endorses the utilization of pre-decision characteristics for equitable and functionally significant assessment of adaptive 6GBO designs.

\begin{table}[htbp]
\caption{All Features Without SMOTE or Threshold Tuning}
\begin{center}

\scriptsize

\begin{tabular}{|l|c|c|c|c|}
\hline
\textbf{Model} & \textbf{Accuracy} & \textbf{F1-Score} & \textbf{ROC-AUC} & \textbf{PR-AUC} \\
\hline

GB & 1.000 & 1.000 & 1.000 & 1.000 \\
\hline

AdaBoost & 1.000 & 1.000 & 1.000 & 1.000 \\
\hline

XGBoost & 1.000 & 1.000 & 1.000 & 1.000 \\
\hline

LightGBM & 1.000 & 1.000 & 1.000 & 1.000 \\
\hline

RF & 0.985 & 0.954 & 1.000 & 1.000 \\
\hline

SVM & 0.905 & 0.732 & 0.960 & 0.862 \\
\hline

LR & 0.885 & 0.716 & 0.940 & 0.812 \\
\hline

KNN & 0.890 & 0.633 & 0.845 & 0.613 \\
\hline

\end{tabular}

\label{tab:all_features_no_smote}

\end{center}
\end{table}

\vspace{-9pt}

\subsection{Explainability using SHAP}

The network FG performed better than other groups: AdaBoost (F1 = 0.327) (Table \ref{tab:default_threshold}); LightGBM (F1 = 0.327) (Table \ref{tab:tuned_threshold}); and SVM (RBF) (F1 = 0.256) (Table \ref{tab:gridsearch_best_models}). However, we employed the AdaBoost model for explainability. To enhance the interpretability of the AdaBoost model for the network FG (F1 = 0.327), analyses of permutation importance (PI), model-specific feature importance (MSFI), and SHAP global importance were conducted. The AdaBoost model performed relatively well in our analysis above. Among these synergistic explainability methods, codebook size, bandwidth, and frequency repeatedly surfaced as the most significant determinants of 6G beamforming enhancement. PI revealed Codebook Size (0.042) and Bandwidth (0.033) as the paramount attributes; however, AdaBoost's MSFI prioritized Codebook Size (0.304), Transmit Power (0.286), and Bandwidth (0.207). Conversely, SHAP analysis for global importance, which measures the contribution of each feature to model predictions, revealed bandwidth (0.073), frequency (0.029), and codebook size (0.020) as the primary influences. The persistent significance of the network feature group through various explainability techniques suggests that codebook arrangement, bandwidth availability, and operating frequency are crucial in influencing 6GBO decisions.

\begin{figure}[t]
    %\centering
    \includegraphics[width=1.0\linewidth]{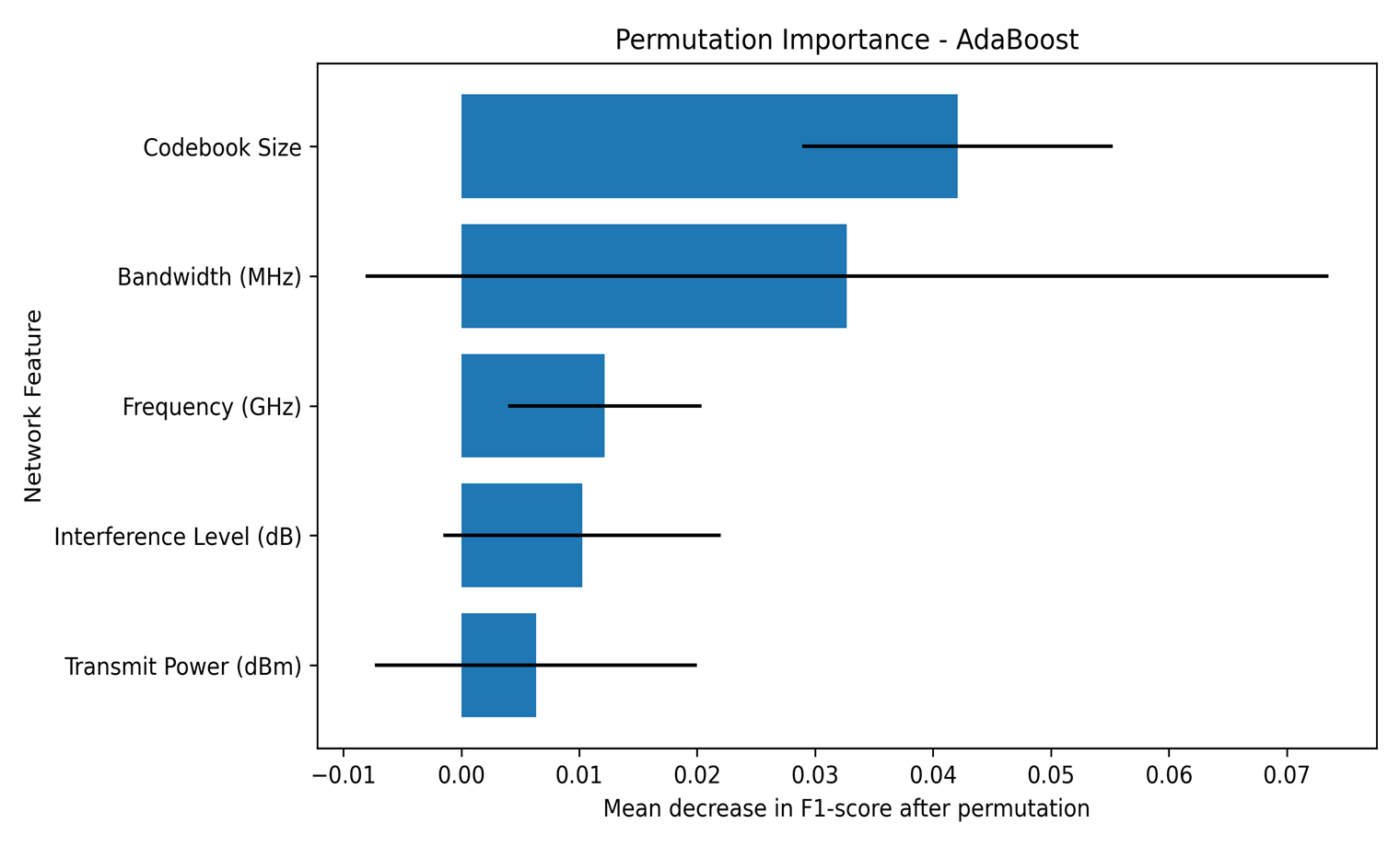}
    \caption{Permutation Importance}
    \label{fig:AdaBoostPI}
\end{figure}

\begin{figure}[t]
    %\centering
    \includegraphics[width=1.0\linewidth]{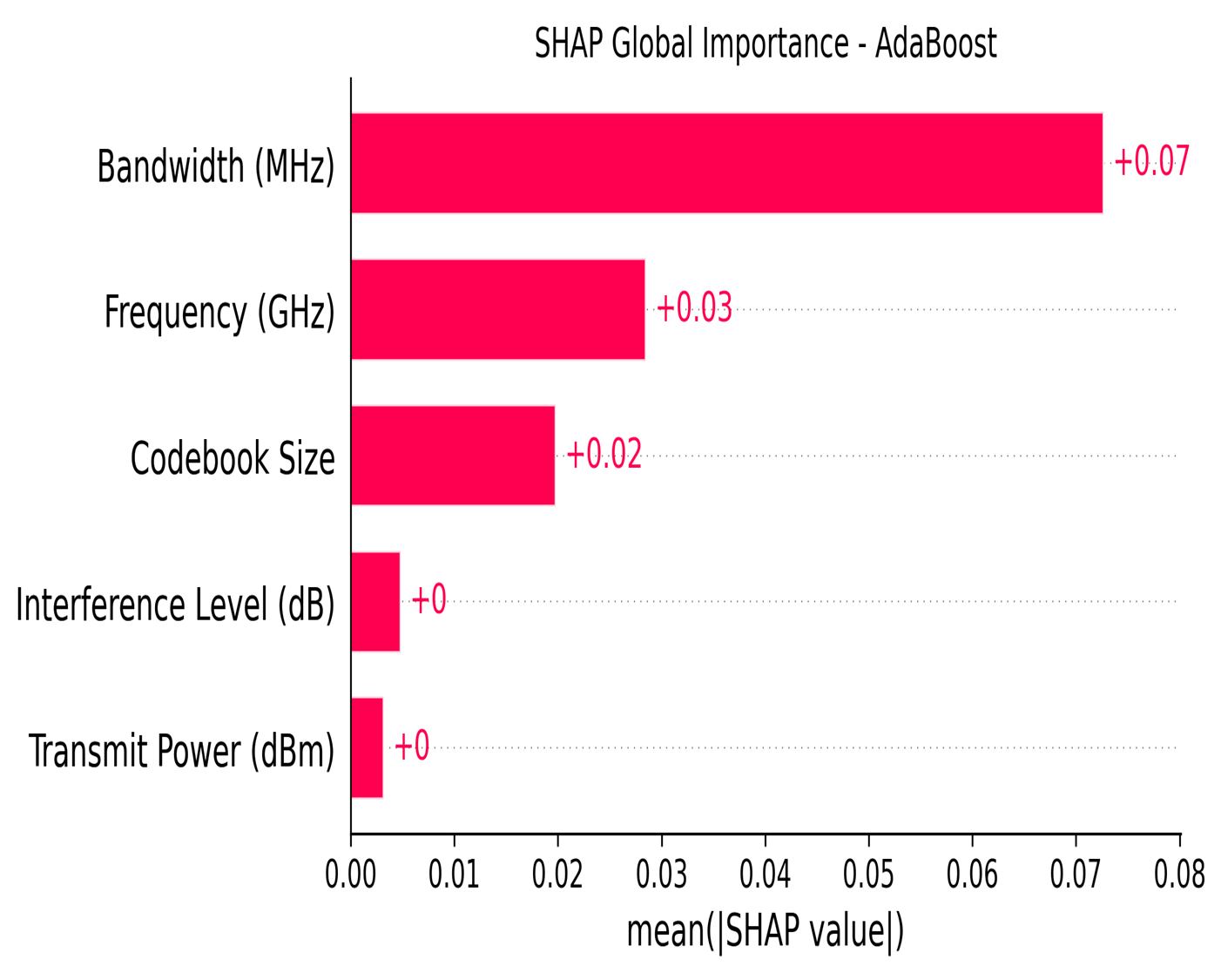}
    \caption{Permutation Importance}
    \label{fig:shapGI}
\end{figure}

\begin{figure}[t]
    %\centering
    \includegraphics[width=1.0\linewidth]{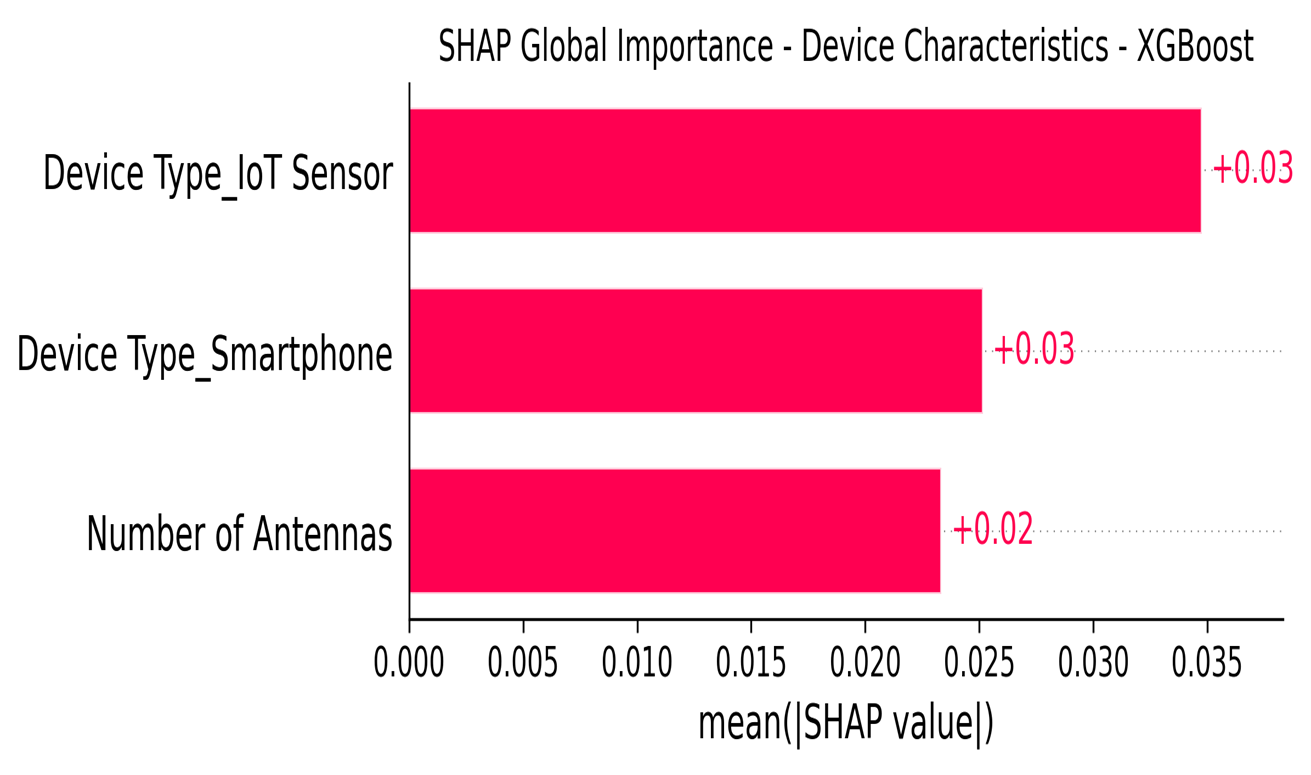}
    \caption{Global Importance for Device Characteristics}
    \label{fig:shapdc}
\end{figure}

\begin{figure}[t]
    %\centering
    \includegraphics[width=1.0\linewidth]{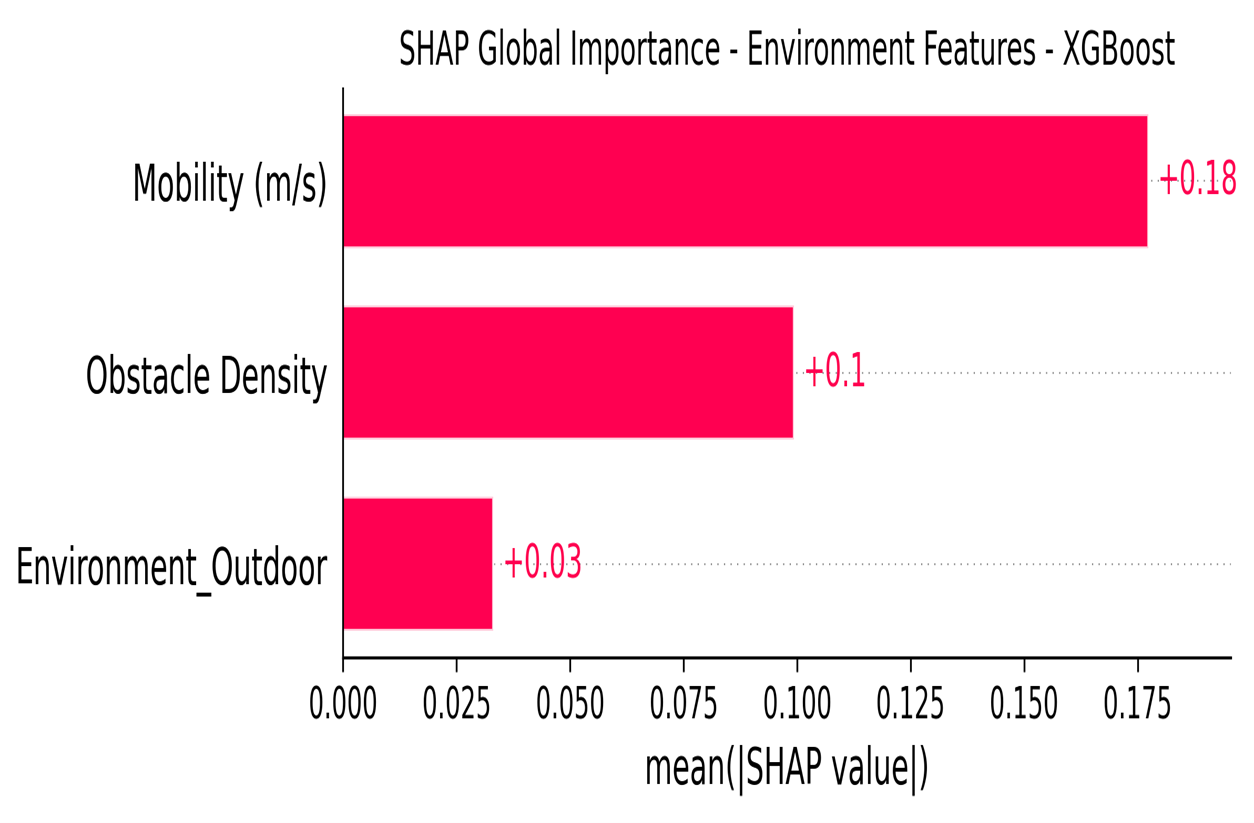}
    \caption{Global Importance for Environmental Factors}
    \label{fig:shapef}
\end{figure}

For completeness, we added  SHAP explainability for the device and environmental attributes. SHAP analysis was conducted to provide insight into the XGBoost models chosen for the DC (F1 = 0.291) and EF (F1 = 0.317) categories. In the DC explainability model (Fig. \ref{fig:shapdc}), the most significant feature was actually Device Type (IoT Sensor) with a SHAP value of 0.035, followed by Device Type (Smartphone) at 0.025 and Number of Antennas at 0.023, suggesting that device type exerts a more substantial influence on 6GBO than the number of antennas. In the EF explainability model (Fig. \ref{fig:shapef}), Mobility was identified as the primary predictor (SHAP = 0.177), succeeded by Obstacle Density (0.099) and Environment Type (Indoor/Outdoor) (0.033). These results indicate that, in terms of XGBoost, user mobility and environmental factors have a more significant impact on device-specific traits than do device-specific traits, with mobility being the paramount environmental element.

\begin{figure}[t]
    %\centering
    \includegraphics[width=1.0\linewidth]{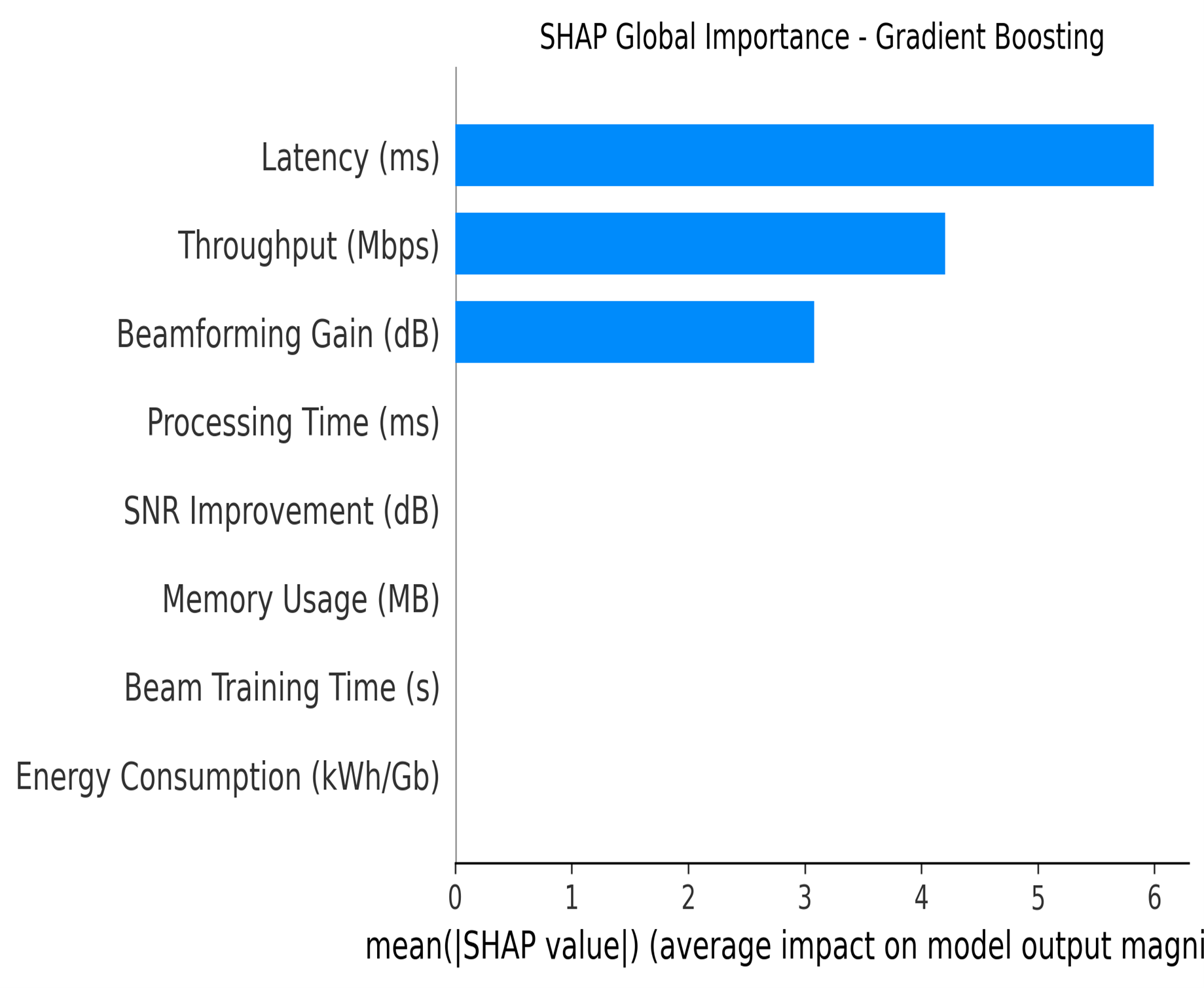}
    \caption{Global Importance for Performance Metrics}
    \label{fig:shappm}
\end{figure}

The GB model utilizing the Model 2 (Performance Metrics) attained flawless classification results. PI, MSFI, and SHAP analysis \ref{fig:shappm} consistently recognized latency, throughput, and beamforming gain as the primary predictors, but the other performance metrics had minimal influence on model decisions. Latency demonstrated the greatest significance among all three techniques, succeeded by Throughput and Beamforming Gain. The nearly flawless prediction accuracy indicates a correlation between these variables and the optimization objective, serving as post-decision performance metrics, which results in target leakage when employed as predictive attributes. As a result, while performance metrics offer significant insights for retrospective analysis and network assessment, they are inadequate for real-time 6GB prediction, underscoring the necessity of relying solely on pre-decision (NP, EF, DC, VA) attributes in practical 6GBO frameworks.

% (Latency, Throughput, Beamforming Gain, Device Type (Smartphone)

\vspace{-4pt}

\subsection{Clustering Network Scenarios}
Here, unsupervised ML clustering approaches were employed to elucidate environmental and operational changes in 6GBO systems with the goal of aiming to identify various profiles of network scenarios. The analysis employed essential contextual characteristics, such as mobility, obstacle density, device type, and environment (outdoor), which jointly define evolving communication contexts and user scenarios. The clustering analysis utilizing performance indicators is designed exclusively for post-hoc (exploratory) study rather than for real-time 6GBO. This is because they can cause target leakage.

\begin{figure}
    \centering
    \includegraphics[width=9cm]{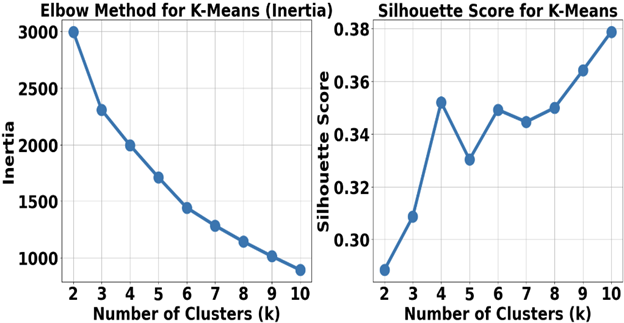}
    \caption{Elbow \& SS Metrics for the Initial Features}
    \label{fig:elbow&silhou}
\end{figure}

We applied three clustering methods—K-means, DBSCAN, and hierarchical clustering—to categorize network scenarios based on feature similarity and spatial density correlations. Prior to clustering, numerical features were standardized, and categorical/binary data were encoded using a preprocessing pipeline. Missing values were addressed with median imputation for numeric data and mode imputation for categorical data. Evaluation metrics, such as the Elbow method and Silhouette score (SS), were used to determine the optimal number of clusters (from 2 to 10 (Table \ref{fig:elbow&silhou})). Furthermore, we employed SelectKBest to identify 5 key performance features for network clustering, focusing on latency, throughput, beamforming gain, device type (smartphone), and SNR improvement to enhance cluster formation despite unsupervised clustering. Table \ref{tab:cluster_importance} contains the feature importance for the initial set (contextual) and selected (performance-based) set.

\begin{figure}
    \centering
    \includegraphics[width=9cm]{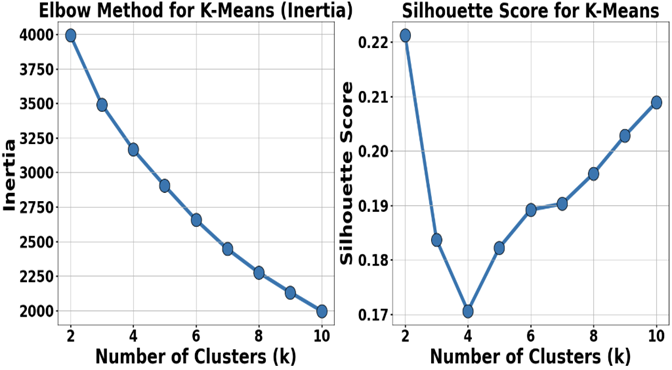}
    \caption{Elbow \& SS Metrics for the Selected Features}
    \label{fig:2Elbow&Sil}
\end{figure}

The ideal number of clusters was calculated separately for each set (initial and selected) with the Elbow and SS techniques. For the initial set of features, k=4 was preferred (Fig. \ref{fig:elbow&silhou}), as it aligned with the elbow point, preserved a commendably high SS, and produced comprehensible operational scenarios. The SS analysis for the SelectKBest-based feature set revealed that k=2 yielded the most pronounced cluster distinctness (Fig. \ref{fig:2Elbow&Sil}); hence, it was used for the unsupervised analysis.

% Principal Component Analysis (PCA) was employed to condense the preprocessed feature space to 2 dimensions for visualization and understanding, facilitating visual illustration of the aforementioned K-Means clusters. The generated labels were subsequently added to the original dataset, and cluster-specific averages of the selected features were computed. This study elucidated the distinctive patterns found in each cluster, including disparities in mobility levels, outdoor deployment conditions, obstacle density, and the existence of IoT sensors, thereby offering insights into various operational contexts associated with the 6GBO dataset.

\begin{figure}
    \centering
    \includegraphics[width=9cm]{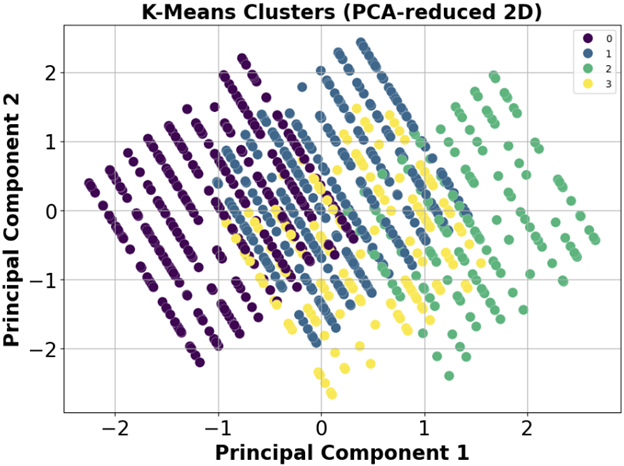}
    \caption{PCA Illustration for the Initial Features}
    \label{fig:pca}
\end{figure}

\begin{figure}
    \centering
    \includegraphics[width=9cm]{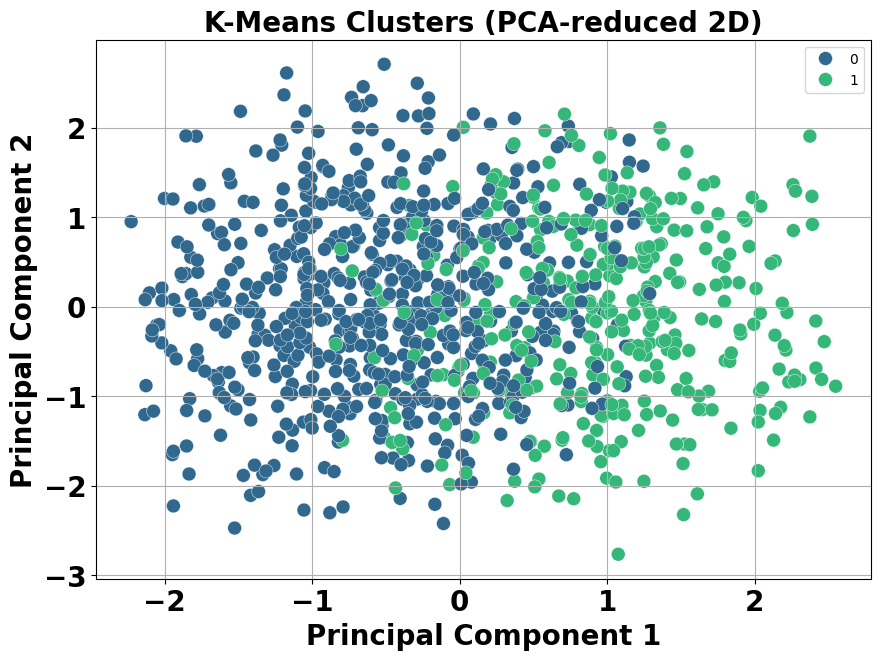}
    \caption{PCA Illustration for the Selected Features}
    \label{fig:2PCA}
\end{figure}

Principal Component Analysis (PCA) was utilized to reduce the preprocessed feature space to two dimensions for better visualization of K-Means clusters (Figs. \ref{fig:pca} and \ref{fig:2PCA}), revealing distinct patterns in contextual features (mobility, deployment conditions, obstacle density, and IoT sensor presence) and performance-related features (latency, throughput, beamforming gain, device type (smartphone), and SNR improvement) within the 6GBO dataset.

Table \ref{tab:cluster_importance} juxtaposes the clustering results of K-Means, DBSCAN, and Hierarchical Clustering, utilizing both feature sets. K-Means and hierarchical clustering demonstrated superior performance for the initial features, attaining an SS of 0.352 at k=4, whereas the SelectKBest-based features yielded a moderately distinguishable 2-cluster arrangement at k=2 with an SS of 0.221. These results suggest that the performance-based features inherently created two overarching behavioral groups, while the original feature set delineates a more intricate 4-cluster framework. With the performance-based parameters, DBSCAN failed to identify any clusters, categorizing each of the 1000 data points as noise. This indicates that density-based clustering may be inappropriate for this dataset or necessitates extensive parameter adjustment.
%Although feature selection altered the feature space, the clustering algorithms typically produced lower SS with the additional characteristics, indicating that the resulting clusters may be less distinct in this modified feature space.

%\vspace{-5pt}

\begin{table}[htbp]
\caption{Feature Importance Comparison for Clustering Analysis}
\label{tab:cluster_importance}
\centering
\scriptsize
\begin{tabular}{|l|c|l|}
\hline
\textbf{Feature} & \textbf{Importance} & \textbf{Feature Set} \\
\hline
Mobility (m/s) & 1.18 & Initial Features \\
Environment\_Outdoor & 0.81 & Initial Features \\
Device Type\_IoT Sensor & 0.56 & Initial Features \\
Obstacle Density & 0.00 & Initial Features \\
\hline
Latency (ms) & 212.27 & Selected Features \\
Throughput (Mbps) & 153.38 & Selected Features \\
Beamforming Gain (dB) & 72.67 & Selected Features \\
Device Type\_Smartphone & 2.33 & Selected Features \\
SNR Improvement (dB) & 1.59 & Selected Features \\
\hline
\end{tabular}
\end{table}

\begin{table}[htbp]
\caption{Clustering Performance Comparison}
\label{tab:clustering_results}
\centering
\scriptsize
\begin{tabular}{|c|l|c|c|c|}
\hline
\textbf{S/No} & \textbf{Algorithm} & \textbf{K} & \textbf{SS} & \textbf{DBI} \\
\hline
0 & K-Means & 4 & 0.352 & 1.320 \\
1 & Hierarchical & 4 & 0.352 & 1.320 \\
2 & DBSCAN & 4 & 0.313 & 1.204 \\
3 & K-Means (Selected Features) & 2 & 0.221 & 1.846 \\
4 & Hierarchical (Selected Features) & 2 &  0.221 & 1.846 \\
%5 & DBSCAN (Tuned eps) & -0.085 & 1.126 \\
\hline
\end{tabular}
\end{table}

In the PCA diagram (Fig. \ref{fig:pca}), context features are delineated distinctly, while in Table \ref{tab:cluster_profiles1}, we summarized clusters against obstacle density (OD), mobility (M), environment\_outdoor (OU), device type\_IoT sensor (DTI), and interpretation. The examination of the cluster centroids identified four unique deployment scenarios, largely distinguished by environmental type and device category. The clusters precisely corresponded to (cluster 0) outdoor non-IoT devices, (cluster 1) indoor non-IoT devices, (cluster 2) indoor IoT sensors, and (cluster 3) outdoor IoT sensors. Conversely, OD and mobility demonstrated minimal fluctuations among clusters, indicating that the clustering pattern was primarily influenced by the installation environment and type of device instead of mobility-based attributes.

Obstacle density, while theoretically pertinent to mmWave obstruction, was uninformative in this dataset. Its dynamic range may be insufficient, inaccurately quantified, or associated with other variables that have already accounted for its impact. Subsequent data collection should either expand its scope or employ more sophisticated blockage proxies, such as LiDAR-derived blockage probabilities or map-driven line of sight. The unsupervised structure within the dataset is predominantly contextual as opposed to performance-oriented.

Segmenting initially by environment and device type produces stable, interpretable cohorts; within each cohort, performance exhibits smooth variation and is more effectively modeled using regression or soft clustering techniques. Context serves as the primary determinant of separable network/user states, whereas KPI variables constitute a continuum that is unable to inherently divide into discrete clusters. This conclusion indicates that contextual considerations may be more significant than motion in differentiating operational beamforming situations within the 6GBO dataset.

\begin{table}[htbp]
\caption{Cluster Profiles Identified from Context Features}
\label{tab:cluster_profiles1}
\centering
\scriptsize
\begin{tabular}{|c|c|c|c|c|p{3.0cm}|}
\hline
\textbf{Cluster} & \textbf{OD} & \textbf{M} & \textbf{OU} & \textbf{DTI} & \textbf{Interpretation} \\
\hline
0 & 5.050 & 1.547 & 1 & 0 & Outdoor, non-IoT Scenario \\
\hline
1 & 4.868 & 1.477 & 0 & 0 & Indoor, non-IoT Scenario \\
\hline
2 & 4.771 & 1.500 & 0 & 1 & Indoor, IoT Sensor Scenario \\
\hline
3 & 4.863 & 1.415 & 1 & 1 & Outdoor, IoT Sensor Scenario \\
\hline
\end{tabular}
\end{table}

Fig. \ref{fig:2PCA} illustrates the PCA visualization of the resultant clusters, demonstrating two moderately distinct groupings with a degree of overlap in the condensed 2-dim space. The identified overlap indicates that the SelectKBest-based features demonstrate gradual shifts between operational states instead of entirely separate classifications. Nonetheless, the clustering reveals that these performance-oriented and device-specific characteristics inherently divide into two overarching behavioral profiles, offering a comprehensible depiction of network operating conditions. This outcome aligns with the SS analysis, which showed k=2 to be the ideal number of clusters for the chosen feature set.

In the PCA diagram (Fig. \ref{fig:2PCA}), the performance-related features are moderately distinguishable using clusters 0 and 1. We observed two moderately distinguishable operating profiles (OPs), and their differences were derived from the above-mentioned selected features (Table \ref{tab:cluster_profiles2}). Cluster 0 shows a non-smartphone profile, defined by reduced latency (5.41 ms), elevated throughput (549.28 Mbps), enhanced beamforming gain (17.80 dB), and superior SNR enhancement (12.69 dB). Conversely, Cluster 1 aligns with a smartphone-centric profile, demonstrating marginally elevated latency (5.47 ms), diminished throughput (526.36 Mbps), worse beamforming gain (17.19 dB), and a lesser SNR enhancement (12.42 dB). Note that the device (smartphone)-type split is clean (0.00 vs. 1.00). These results suggest that the chosen performance and device-specific attributes inherently divide the 6G network into two OPs with modest differences, offering a comprehensible post-hoc analysis of beamforming optimization performance.

\begin{table}[!t]
\caption{Clustering Profiles from Performance-based Features}
\label{tab:cluster_profiles2}
\centering
\scriptsize
\begin{tabular}{|l|c|c|c|}
\hline
\textbf{Selected Features} & \textbf{Cluster 0} & \textbf{Cluster 1}   & \textbf{Difference} \\
\hline
Latency (ms)              & 5.41   & 5.47 & -0.06 \\
Throughput (Mbps)         & 549.28  & 526.36 & +22.92 \\
Beamforming Gain (dB)     & 17.80   & 17.19 & +0.61 \\
Device Type (Smartphone) (0/1)  & 0.00   & 1.00  & — \\
SNR Improvement (dB)      & 12.69  & 12.42 & +0.27 \\
\hline
\end{tabular}
\end{table}

Finally, it is noteworthy that TT significantly enhanced recall, with multiple models attaining values near 1.000, signifying that almost all 6GBO prospects were identified. Nonetheless, this enhancement was achieved at the cost of accuracy, which remained comparatively low, around 0.17–0.19. This indicates that numerous samples identified as needing optimization were incorrect positives, leading to an elevated false-alarm rate. In adaptive beamforming, such erroneous detections could lead the network to inappropriately initiate beam reconfiguration and training, codebook searches, or other optimization processes for links that do not need modification or adjustment \cite{giordani2018tutorial}. These additional beam updates may elevate computational demands, energy usage, signaling traffic, and processing delays, thereby diminishing overall system efficiency \cite{giordani2018tutorial}. In extremely dynamic 6G settings, emphasizing recall could be advantageous when the expense of overlooking a legitimate 6GBO opportunity surpasses that of executing an unwarranted update. A false negative can result in diminished link quality, lowered throughput, and heightened communication latency. Thus, the criterion for decision-making ought to be contingent upon the application and must weigh the operational expenses of superfluous beam adjustments against the potential for missed 6GBO prospects.

\section{Conclusion}
\label{con}

This paper addresses practical, systems-level questions that are of interest to researchers and practitioners working on edge intelligence and beam management in mmWave/6G deployments. Specifically, the study presents a systematic ML study of beamforming optimization for 6G-IoT, comparing the predictive power of network, environmental, and device feature groups and demonstrating how ML pipelines (with explainability and imbalance handling) can guide adaptive, low-latency beam management. Generally, the network parameters have stronger F1-score and ROC-AUC/PR-AUC values. However, the device attributes achieved the strongest F1-score when GridSearchCV was employed. Despite the substantial enhancement in detecting optimized 6G beamforming scenarios through threshold tuning, the resultant rise in false positives underscores the necessity for tailored threshold preference that reconciles detection efficacy with operational demands. In the future, we would compare pre-decision feature groups against a simple majority-class baseline and a calibrated probability baseline. Also, our comparative analysis would extend to predicting beamforming gain, latency, energy consumption, throughput, and SNR improvement using deep and reinforcement learning techniques.

%ACKNOWLEDGEMENT 

\section{Acknowledgment}
This work was undertaken thanks in part to funding from the Connected Minds Program, supported by the Canada First Research Excellence Fund, Grant \#CFREF-2022-00010.

\bibliographystyle{ieeetr}
\bibliography{references}

@misc{ziya2024,
  author       = {Ziya, M.},
  title        = {6G IoT Intelligent Management Dataset: Computer Vision-Assisted Intelligent Beamforming Management in 6G Networks},
  year         = {2024},
  howpublished = {\url{https://www.kaggle.com/datasets/ziya07/6g-iot-intelligent-management-dataset}},
  note         = {Accessed: 2025-10-21},
}

@article{giordani2020toward,
  title={Toward 6G networks: Use cases and technologies},
  author={Giordani, Marco and Polese, Michele and Mezzavilla, Marco and Rangan, Sundeep and Zorzi, Michele},
  journal={IEEE communications magazine},
  volume={58},
  number={3},
  pages={55--61},
  year={2020},
  publisher={IEEE}
}

@article{huang2018deep,
  title={Deep learning for super-resolution channel estimation and DOA estimation based massive MIMO system},
  author={Huang, Hongji and Yang, Jie and Huang, Hao and Song, Yiwei and Gui, Guan},
  journal={IEEE Transactions on Vehicular Technology},
  volume={67},
  number={9},
  pages={8549--8560},
  year={2018},
  publisher={IEEE}
}

@article{qin2019deep,
  title={Deep learning in physical layer communications},
  author={Qin, Zhijin and Ye, Hao and Li, Geoffrey Ye and Juang, Biing-Hwang Fred},
  journal={IEEE Wireless Communications},
  volume={26},
  number={2},
  pages={93--99},
  year={2019},
  publisher={IEEE}
}

@article{saad2019vision,
  title={A vision of 6G wireless systems: Applications, trends, technologies, and open research problems},
  author={Saad, Walid and Bennis, Mehdi and Chen, Mingzhe},
  journal={IEEE network},
  volume={34},
  number={3},
  pages={134--142},
  year={2019},
  publisher={IEEE}
}

@article{ye2019deep,
  title={Deep reinforcement learning based resource allocation for V2V communications},
  author={Ye, Hao and Li, Geoffrey Ye and Juang, Biing-Hwang Fred},
  journal={IEEE Transactions on Vehicular Technology},
  volume={68},
  number={4},
  pages={3163--3173},
  year={2019},
  publisher={IEEE}
}

@article{AKBAR2025104040,
title = {On challenges of sixth-generation (6G) wireless networks: A comprehensive survey of requirements, applications, and security issues},
journal = {Journal of Network and Computer Applications},
volume = {233},
pages = {104040},
year = {2025},
issn = {1084-8045},
doi = {https://doi.org/10.1016/j.jnca.2024.104040},
url = {https://www.sciencedirect.com/science/article/pii/S1084804524002170},
author = {Muhammad Sajjad Akbar and Zawar Hussain and Muhammad Ikram and Quan Z. Sheng and Subhas Chandra Mukhopadhyay}
}

@article{QADIR2023296,
title = {Towards 6G Internet of Things: Recent advances, use cases, and open challenges},
journal = {ICT Express},
volume = {9},
number = {3},
pages = {296-312},
year = {2023},
issn = {2405-9595},
doi = {https://doi.org/10.1016/j.icte.2022.06.006},
url = {https://www.sciencedirect.com/science/article/pii/S2405959522000959},
author = {Zakria Qadir and Khoa N. Le and Nasir Saeed and Hafiz Suliman Munawar}
}

@article{elbir2023twenty,
  title={Twenty-five years of advances in beamforming: From convex and nonconvex optimization to learning techniques},
  author={Elbir, Ahmet M and Mishra, Kumar Vijay and Vorobyov, Sergiy A and Heath, Robert W},
  journal={IEEE Signal Processing Magazine},
  volume={40},
  number={4},
  pages={118--131},
  year={2023},
  publisher={IEEE}
}

@article{brilhante2023literature,
  title={A literature survey on AI-aided beamforming and beam management for 5G and 6G systems},
  author={Brilhante, Davi da Silva and Manjarres, Joanna Carolina and Moreira, Rodrigo and de Oliveira Veiga, Lucas and de Rezende, Jos{\'e} F and M{\"u}ller, Francisco and Klautau, Aldebaro and Leonel Mendes, Luciano and P. de Figueiredo, Felipe A},
  journal={Sensors},
  volume={23},
  number={9},
  pages={4359},
  year={2023},
  publisher={MDPI}
}

@book{duong2023ultra,
  title={Ultra-reliable and Low-Latency Communications (URLLC) theory and practice: Advances in 5G and beyond},
  author={Duong, Trung Q and Khosravirad, Saeed R and She, Changyang and Popovski, Petar and Bennis, Mehdi and Quek, Tony QS},
  year={2023},
  publisher={John Wiley \& Sons}
}

@article{heath2016overview,
  title={An overview of signal processing techniques for millimeter wave MIMO systems},
  author={Heath, Robert W and Gonzalez-Prelcic, Nuria and Rangan, Sundeep and Roh, Wonil and Sayeed, Akbar M},
  journal={IEEE journal of selected topics in signal processing},
  volume={10},
  number={3},
  pages={436--453},
  year={2016},
  publisher={IEEE}
}

@article{ZHOU2024,
title = {A comprehensive survey of artificial intelligence applications in UAV-enabled wireless networks},
journal = {Digital Communications and Networks},
volume = {12},
number = {4},
pages = {561-583},
year = {2026},
issn = {2352-8648},
doi = {https://doi.org/10.1016/j.dcan.2024.11.005},
url = {https://www.sciencedirect.com/science/article/pii/S2352864824001536},
author = {Li Zhou and Hao Yin and Haitao Zhao and Jibo Wei and Dewen Hu and Victor C.M. Leung}
}

@Inbook{Benesty2021,
author="Benesty, Jacob
and Cohen, Israel
and Chen, Jingdong",
title="A Brief Overview of Conventional Beamforming",
bookTitle="Array Beamforming with Linear Difference Equations",
year="2021",
publisher="Springer International Publishing",
address="Cham",
pages="13--21",
isbn="978-3-030-68273-6",
doi="10.1007/978-3-030-68273-6_2",
url="https://doi.org/10.1007/978-3-030-68273-6_2"
}

@misc{sharif_2025,
  author       = {Raad Sharar Sharif},
  title        = {Optimized Beamforming 6G XAI},
  year         = {2025},
  howpublished = {\url{https://github.com/raadsr15/Optimized-Beamforming-6G-XAI}},
  note         = {Accessed: Jul. 20, 2026}
}

@INPROCEEDINGS{Klautau2018,
  author={Klautau, Aldebaro and Batista, Pedro and González-Prelcic, Nuria and Wang, Yuyang and Heath, Robert W.},
  booktitle={2018 Information Theory and Applications Workshop (ITA)}, 
  title={5G MIMO Data for Machine Learning: Application to Beam-Selection Using Deep Learning}, 
  year={2018},
  volume={},
  number={},
  pages={1-9},
  doi={10.1109/ITA.2018.8503086}}

@article{giordani2018tutorial,
  title={A tutorial on beam management for 3GPP NR at mmWave frequencies},
  author={Giordani, Marco and Polese, Michele and Roy, Arnab and Castor, Douglas and Zorzi, Michele},
  journal={IEEE Communications Surveys \& Tutorials},
  volume={21},
  number={1},
  pages={173--196},
  year={2018},
  publisher={IEEE}
}

\vspace{12pt}
\end{document}